\documentclass{wscpaperproc}
\usepackage{latexsym}
\usepackage{graphicx}
\usepackage{mathptmx}
\usepackage[T1]{fontenc}

\usepackage{amsmath}
\usepackage{amsfonts}
\usepackage{amssymb}
\usepackage{amsbsy}
\usepackage{amsthm}
\usepackage{algorithm, algpseudocode}
\usepackage{graphicx, subcaption}
\usepackage{multirow}

\usepackage[pdftex,colorlinks=true,urlcolor=blue,citecolor=black,anchorcolor=black,linkcolor=black]{hyperref}

\newtheoremstyle{wsc}
{3pt}
{3pt}
{}
{}
{\bf}
{}
{.5em}
{}

\theoremstyle{wsc}
\newtheorem{theorem}{Theorem}

\newtheorem{proposition}[theorem]{Proposition}

\begin{document}
\def\noi{\noindent}
\def\ba{{\mathbf{a}}}
\def\bb{{\mathbf{b}}}
\def\bc{{\mathbf{c}}}
\def\bd{\mathbf{d}}
\def\be{\mathbf{e}}
\def\bff{\mathbf{f}}
\def\bg{\mathbf{g}}
\def\bh{\mathbf{h}}
\def\bi{\mathbf{i}}
\def\bj{\mathbf{j}}
\def\bk{\mathbf{k}}
\def\bl{\mathbf{l}}
\def\bm{\mathbf{m}}
\def\bn{\mathbf{n}}
\def\bp{\mathbf{p}}
\def\bq{\mathbf{q}}
\def\br{\mathbf{r}}
\def\bs{\mathbf{s}}
\def\bt{\mathbf{t}}
\def\bu{\mathbf{u}}
\def\bv{\mathbf{v}}
\def\bw{\mathbf{w}}
\def\bx{\mathbf{x}}  
\def\by{\mathbf{y}}
\def\bz{\mathbf{z}}
\def\bA{\mathbf{A}}
\def\bB{\mathbf{B}}
\def\bC{\mathbf{C}}
\def\bD{\mathbf{D}}
\def\bE{\mathbf{E}}
\def\bF{\mathbf{F}}
\def\bG{\mathbf{G}}
\def\bH{\mathbf{H}}
\def\bI{\mathbf{I}}
\def\bJ{\mathbf{J}}
\def\bK{\mathbf{K}}
\def\bL{\mathbf{L}}
\def\bM{\mathbf{M}}
\def\bN{\mathbf{N}}
\def\bO{\mathbf{O}}
\def\bP{\mathbf{P}}
\def\bQ{\mathbf{Q}}
\def\bR{\mathbf{R}}
\def\bS{\mathbf{S}}
\def\bU{\mathbf{U}}
\def\bT{\mathbf{T}}
\def\bY{\mathbf{Y}}
\def\bX{\mathbf{X}}
\def\bV{\mathbf{V}}
\def\bW{\mathbf{W}}
\def\bZ{\mathbf{Z}}
\def\bo{\mathbf{0}}

\def\bfepsilon{{\boldsymbol{\epsilon}}}
\def\bflambda{{\boldsymbol{\lambda}}}
\def\bfgamma{{\boldsymbol{\gamma}}}
\def\bfGamma{{\boldsymbol{\Gamma}}}
\def\bfLambda{{\boldsymbol{\Lambda}}}
\def\bfOmega{{\boldsymbol{\Omega}}}
\def\bfDelta{{\boldsymbol{\Delta}}}
\def\bfalpha{{\boldsymbol{\alpha}}}
\def\bfmu{{\boldsymbol{\mu}}}
\def\bfnu{{\boldsymbol{\nu}}}
\def\bfSigma{{\boldsymbol{\Sigma}}}
\def\bfsigma{{\boldsymbol{\sigma}}}
\def\bfpi{{\boldsymbol{\pi}}}
\def\bfPi{{\boldsymbol{\Pi}}}
\def\bftheta{{\boldsymbol{\theta}}}
\def\bfTheta{{\boldsymbol{\Theta}}}
\def\bfxi{{\boldsymbol{\xi}}}
\def\bfbeta{{\boldsymbol{\beta}}}
\def\bfrho{{\boldsymbol{\rho}}}
\def\bfzeta{{\boldsymbol{\zeta}}}
\def\bfphi{{\boldsymbol{\phi}}}
\def\bfpsi{{\boldsymbol{\psi}}}
\def\bfPsi{{\boldsymbol{\Psi}}}

\def\cA{\mathcal{A}}
\def\cB{\mathcal{B}}
\def\cC{\mathcal{C}}
\def\cD{\mathcal{D}}
\def\cE{\mathcal{E}}
\def\cF{\mathcal{F}}
\def\cG{\mathcal{G}}
\def\cH{\mathcal{H}}
\def\cI{\mathcal{I}}
\def\cJ{\mathcal{J}}
\def\cK{\mathcal{K}}
\def\cL{\mathcal{L}}
\def\cM{\mathcal{M}}
\def\cN{\mathcal{N}}
\def\cO{\mathcal{O}}
\def\cP{\mathcal{P}}
\def\cQ{\mathcal{Q}}
\def\cR{\mathcal{R}}
\def\cS{\mathcal{S}}
\def\cT{\mathcal{T}}
\def\cU{\mathcal{U}}
\def\cV{\mathcal{V}}
\def\cW{\mathcal{W}}
\def\cX{\mathcal{X}}
\def\cY{\mathcal{Y}}
\def\cZ{\mathcal{Z}}

\def\mA{\mathbb{A}}
\def\mB{\mathbb{B}}
\def\mC{\mathbb{C}}
\def\mD{\mathbb{D}}
\def\mE{\mathbb{E}}
\def\mF{\mathbb{F}}
\def\mG{\mathbb{G}}
\def\mH{\mathbb{H}}
\def\mI{\mathbb{I}}
\def\mJ{\mathbb{J}}
\def\mK{\mathbb{K}}
\def\mL{\mathbb{L}}
\def\mM{\mathbb{M}}
\def\mN{\mathbb{N}}
\def\mO{\mathbb{O}}
\def\mP{\mathbb{P}}
\def\mQ{\mathbb{Q}}
\def\mR{\mathbb{R}}
\def\mS{\mathbb{S}}
\def\mT{\mathbb{T}}
\def\mU{\mathbb{U}}
\def\mV{\mathbb{V}}
\def\mW{\mathbb{W}}
\def\mX{\mathbb{X}}
\def\mY{\mathbb{Y}}
\def\mZ{\mathbb{Z}}

\def\tA{{\tilde{A}}}
\def\tS{{\tilde{S}}}
\def\tR{{\tilde{R}}}
\def\tE{{\tilde{E}}}
\def\ts{{\tilde{s}}}
\def\tX{{\tilde{X}}}
\def\tpi{{\tilde{\pi}}}

\def\eq{\[}
\def\en{\]}

\def\smskip{\smallskip}
\def\sep#1{ \hskip -5pt & #1 & \hskip -5 pt}
\def\texitem#1{\par\smskip\noindent\hangindent 25pt
               \hbox to 25pt {\hss #1 ~}\ignorespaces}

\def\abs#1{|#1|}
\def\norm#1{\|#1\|}
\def\vec#1{\left(\!\begin{array}{c}#1\end{array}\!\right)}
\def\bmat#1{\left[\begin{array}{cc}#1\end{array}\right]}

\newcommand{\BEAS}{\begin{eqnarray*}}
\newcommand{\EEAS}{\end{eqnarray*}}
\newcommand{\BEA}{\begin{eqnarray}}
\newcommand{\EEA}{\end{eqnarray}}
\newcommand{\BEQ}{\begin{eqnarray}}
\newcommand{\EEQ}{\end{eqnarray}}
\newcommand{\BIT}{\begin{itemize}}
\newcommand{\EIT}{\end{itemize}}
\newcommand{\BNUM}{\begin{enumerate}}
\newcommand{\ENUM}{\end{enumerate}}

\newcommand{\BA}{\begin{array}}
\newcommand{\EA}{\end{array}}

\newcommand{\cf}{{\it cf.}}
\newcommand{\eg}{{\it e.g.}}
\newcommand{\ie}{{\it i.e.}}
\newcommand{\etc}{{\it etc.}}

\newcommand{\ones}{\mathbf 1}

\newcommand{\reals}{\mathbb{R}}
\newcommand{\integers}{\mathbb{Z}}
\newcommand{\eqbydef}{\mathrel{\stackrel{\Delta}{=}}}
\newcommand{\complex}{{\mbox{\bf C}}}


\newcommand{\Span}{\mbox{\textrm{span}}}
\newcommand{\Range}{\mbox{\textrm{range}}}
\newcommand{\nullspace}{{\mathcal N}}
\newcommand{\range}{{\mathcal R}}
\newcommand{\Nullspace}{\mbox{\textrm{nullspace}}}
\newcommand{\Rank}{\mathop{\bf rank}}
\newcommand{\Tr}{\mathop{\bf Tr}}
\newcommand{\detr}{\mathop{\bf det}}
\newcommand{\diag}{\mathop{\bf diag}}
\newcommand{\lambdamax}{{\lambda_{\rm max}}}
\newcommand{\lambdamin}{\lambda_{\rm min}}

\newcommand{\Expect}{\mathop{\bf E{}}}
\newcommand{\Prob}{\mathop{\bf Prob}}
\newcommand{\erf}{\mathop{\bf erf}}
\newcommand{\iid}{\overset{iid}{\sim}}

\def\co#1{{\mathop {\bf cone}}\left(#1\right)}
\def\conv#1{\mathop {\bf conv}\left(#1\right)}
\def\var#1{\mathop{\bf Var}\big[#1\big]}
\newcommand{\dist}{\mathop{\bf dist{}}}
\newcommand{\Ltwo}{{\bf L}_2}

\newcommand{\argmin}{\mathop{\rm argmin}}
\newcommand{\argmax}{\mathop{\rm argmax}}
\newcommand{\epi}{\mathop{\bf epi}}
\newcommand{\hypo}{\mathop{\bf hypo}}
\newcommand{\ri}{\mathop{\bf ri}}
\newcommand{\vol}{\mathop{\bf vol}}
\newcommand{\Vol}{\mathop{\bf vol}}

\newcommand{\dom}{\mathop{\bf dom}}
\newcommand{\aff}{\mathop{\bf aff}}
\newcommand{\cl}{\mathop{\bf cl}}
\newcommand{\Angle}{\mathop{\bf angle}}
\newcommand{\intr}{\mathop{\bf int}}
\newcommand{\relint}{\mathop{\bf rel int}}
\newcommand{\bdry}{\mathop{\bf bd}}
\newcommand{\vect}{\mathop{\bf vec}}
\newcommand{\dsp}{\displaystyle}
\newcommand{\foequal}{\simeq}
\newcommand{\VOL}{{\mbox{\bf vol}}}

\newcommand{\xopt}{x^{\rm opt}}
\newcommand{\Xb}{{\mbox{\bf X}}}
\newcommand{\xst}{x^\star}
\newcommand{\varphist}{\varphi^\star}
\newcommand{\lambdast}{\lambda^\star}
\newcommand{\Zst}{Z^\star}
\newcommand{\fstar}{f^\star}
\newcommand{\xstar}{x^\star}
\newcommand{\xc}{x^\star}
\newcommand{\lambdac}{\lambda^\star}
\newcommand{\lambdaopt}{\lambda^{\rm opt}}

\newcommand{\geqK}{\mathrel{\succeq_K}}
\newcommand{\gK}{\mathrel{\succ_K}}
\newcommand{\leqK}{\mathrel{\preceq_K}}
\newcommand{\lK}{\mathrel{\prec_K}}
\newcommand{\geqKst}{\mathrel{\succeq_{K^*}}}
\newcommand{\gKst}{\mathrel{\succ_{K^*}}}
\newcommand{\leqKst}{\mathrel{\preceq_{K^*}}}
\newcommand{\lKst}{\mathrel{\prec_{K^*}}}
\newcommand{\geqL}{\mathrel{\succeq_L}}
\newcommand{\gL}{\mathrel{\succ_L}}
\newcommand{\leqL}{\mathrel{\preceq_L}}
\newcommand{\lL}{\mathrel{\prec_L}}
\newcommand{\geqLst}{\mathrel{\succeq_{L^*}}}
\newcommand{\gLst}{\mathrel{\succ_{L^*}}}
\newcommand{\leqLst}{\mathrel{\preceq_{L^*}}}
\newcommand{\lLst}{\mathrel{\prec_{L^*}}}
\def\join{\vee}
\def\meet{\wedge}

\newcommand{\Var}{\mathrm{Var}}
\newcommand{\Corr}{\mathrm{Corr}}
\newcommand{\Cov}{\mathrm{Cov}}

\def\red#1{\textcolor{red}{#1}}
\def\blue#1{\textcolor{blue}{#1}}
\def\green#1{{\textcolor{OliveGreen}{#1}}}


\newcommand{\EE}{\mathbb{E}}
\newcommand{\PP}{\mathbb{P}}

\newcommand{\rp}{\overset{P}{\rightarrow} }
\newcommand{\rd}{\overset{D}{\rightarrow} }
\newcommand{\ras}{\overset{a.s.}{\rightarrow}}
\newcommand{\rlp}{\overset{L_{p}}{\rightarrow} }
%
%

\pagestyle{fancyplain}

\thispagestyle{plain}
\firstPageHead{}

\chead{\fancyplain{}{\itshape Zhao, and Chen}}

\rhead{}
\cfoot{}
\renewcommand{\headrulewidth}{0pt} 

\makeatletter
\let\@internalcite\cite
\def\cite{\def\@citeseppen{-1000}%
    \def\@cite##1##2{(##1\if@tempswa , ##2\fi)}%
    \def\citeauthoryear##1##2##3{##1 ##3}\@internalcite}
\def\citeNP{\def\@citeseppen{-1000}%
    \def\@cite##1##2{##1\if@tempswa , ##2\fi}%
    \def\citeauthoryear##1##2##3{##1 ##3}\@internalcite}
\def\citeN{\def\@citeseppen{-1000}%
    \def\@cite##1##2{##1\if@tempswa, ##2)\else{}\fi}%
    \def\citeauthoryear##1##2##3{##1 (##3)}\@citedata}
\def\citeA{\def\@citeseppen{-1000}%
    \def\@cite##1##2{(##1\if@tempswa , ##2\fi)}%
    \def\citeauthoryear##1##2##3{##1}\@internalcite}
\def\citeANP{\def\@citeseppen{-1000}%
    \def\@cite##1##2{##1\if@tempswa , ##2\fi}%
    \def\citeauthoryear##1##2##3{##1}\@internalcite}
\def\shortcite{\def\@citeseppen{-1000}%
    \def\@cite##1##2{(##1\if@tempswa , ##2\fi)}%
    \def\citeauthoryear##1##2##3{##2 ##3}\@internalcite}
\def\shortciteNP{\def\@citeseppen{-1000}%
    \def\@cite##1##2{##1\if@tempswa , ##2\fi}%
    \def\citeauthoryear##1##2##3{##2 ##3}\@internalcite}
\def\shortciteN{\def\@citeseppen{-1000}%
    \def\@cite##1##2{##1\if@tempswa, ##2\else{}\fi}%
    \def\citeauthoryear##1##2##3{##2 (##3)}\@citedata}
\def\shortciteA{\def\@citeseppen{-1000}%
    \def\@cite##1##2{(##1\if@tempswa , ##2\fi)}%
    \def\citeauthoryear##1##2##3{##2}\@internalcite}
\def\shortciteANP{\def\@citeseppen{-1000}%
    \def\@cite##1##2{##1\if@tempswa , ##2\fi}%
    \def\citeauthoryear##1##2##3{##2}\@internalcite}
\def\citeyear{\def\@citeseppen{-1000}%
    \def\@cite##1##2{(##1\if@tempswa , ##2\fi)}%
    \def\citeauthoryear##1##2##3{##3}\@citedata}
\def\citeyearNP{\def\@citeseppen{-1000}%
    \def\@cite##1##2{##1\if@tempswa , ##2\fi}%
    \def\citeauthoryear##1##2##3{##3}\@citedata}
%
%
%
\def\@citedata{%
    \@ifnextchar [{\@tempswatrue\@citedatax}%
                  {\@tempswafalse\@citedatax[]}%
}

\def\@citedatax[#1]#2{%
\if@filesw\immediate\write\@auxout{\string\citation{#2}}\fi%
  \def\@citea{}\@cite{\@for\@citeb:=#2\do%
    {\@citea\def\@citea{, }\@ifundefined
       {b@\@citeb}{{\bf ?}%
       \@warning{Citation `\@citeb' on page \thepage \space undefined}}%
{\csname b@\@citeb\endcsname}}}{#1}}%

%
\def\@citex[#1]#2{%
\if@filesw\immediate\write\@auxout{\string\citation{#2}}\fi%
  \def\@citea{}\@cite{\@for\@citeb:=#2\do%
    {\@citea\def\@citea{; }\@ifundefined
       {b@\@citeb}{{\bf ?}%
       \@warning{Citation `\@citeb' on page \thepage \space undefined}}%
{\csname b@\@citeb\endcsname}}}{#1}}%

%
\def\@biblabel#1{}
\makeatother



\newdimen\bibindent
\bibindent=0.0em
\def\thebibliography#1{\section*{\refname}\list
   {}{\settowidth\labelwidth{[#1]}
   \leftmargin\parindent
   \itemindent -\parindent
   \listparindent \itemindent
   \itemsep 0pt
   \parsep 0pt}
   \def\newblock{}
   \sloppy
   \sfcode`\.=1000\relax}


\setlength{\baselineskip}{12.7pt}

\title{When Machine Learning Meets Importance Sampling: A More Efficient Rare Event Estimation Approach}

\author{\begin{center}Ruoning Zhao\textsuperscript{1}, and Xinyun Chen\textsuperscript{1}\\
[11pt]
\textsuperscript{1}School of Data Science, the Chinese University of Hong Kong, Shenzhen, GUANGDONG, CHINA\end{center}
}

\maketitle
\vspace{-12pt}

\section*{ABSTRACT}
	Driven by applications in telecommunication networks, we explore the simulation task of estimating rare event probabilities for tandem queues in their steady state. Existing literature has recognized that importance sampling methods can be inefficient, due to the exploding variance of the path-dependent likelihood functions. To mitigate this, we introduce a new importance sampling approach that utilizes a marginal likelihood ratio on the stationary distribution, effectively avoiding the issue of excessive variance. In addition, we design a machine learning algorithm to estimate this marginal likelihood ratio using importance sampling data. Numerical experiments indicate that our algorithm outperforms the classic importance sampling methods.

\section{INTRODUCTION}
\label{sec:intro}
In computing and telecommunication industries, service level agreements (SLA) in commercial contracts are aimed to ensure that the promised quality of service (QoS) is met under a given network design. A commonly used QoS performance metric in SLA takes the form of a guarantee on the tail probability of certain queueing function of the network in steady state 
	\cite{milner2008service},  i.e.,
	\begin{align*}
		\mP(X_{\infty} > \gamma)\leq p,
	\end{align*}
	where the random variable $X_{\infty}$ could be the steady-state total number of jobs in the system or a job's sojourn time, and $p$ is a small number. Intuitively, this type of SLA aims to maintain the network congestion within an acceptable threshold with a high probability in the long run. Since obtaining an analytic expression for the stationary distribution is often not feasible for general stochastic network models, the industry generally relies on simulation techniques to numerically compute the probability $\mP(X_{\infty} > \gamma)$, thereby assessing the SLA.

	Since the value of $p$ is typically very small, like $p = 10^{-5}$ \cite{harchol2021open}, calculating the probability $\mP(X_{\infty} > \gamma)$ essentially becomes a problem of rare event simulation. A common approach to improve the efficiency of rare event simulation is by importance sampling (IS). Nonetheless, estimating the stationary distribution $X_\infty$ adds an additional level of complexity to the design of the IS algorithms. Earlier studies \cite{glasserman1995analysis,de2006analysis}  have demonstrated that finding an importance distribution that could lead to significant variance reduction for estimating tail probabilities, even in a basic two-node tandem queueing network, is quite challenging.
	
	In this paper, we highlight that these challenges largely originate from the path-dependent characteristics of the classic  importance sampling algorithms, which use the regenerative structures of queueing processes to deal with the stationary distribution. To tackle this issue, we introduce a novel IS algorithm, demonstrated through the two-node tandem queueing network example. This algorithm applies importance sampling directly on the "marginal" stationary distribution of individual samples in the state space, rather than on the regenerative cycle sample paths. Given that the marginal likelihood ratio of stationary distributions is in general unknown for queueing networks, we use the off-policy evaluation method from reinforcement learning literature to approximate the stationary likelihood ratio from simulation data. Broadly speaking, our approach integrates traditional importance sampling with machine learning techniques. The numerical findings indicate that our algorithm outperforms the traditional regenerative importance sampling method, achieving lower mean squared errors. Furthermore, the advantage of our algorithm is relatively robust with respect to the choice of importance distribution.
	
	The rest of the paper is organized as follows. We provide a brief literature review on related topics in Section 2. In Section 3, we introduce the estimation objective in the tandem queues and discuss the drawbacks of the state-of-the-art IS method. In Section 4, we first explain the key components in algorithm design and then present the complete algorithm.  Numerical results are reported in Section 5. Section 6 concludes the paper with insights into future research directions.

\section{RELATED WORKS}
The rare event simulation in queueing systems, like the response time violations, is typically addressed by importance sampling methods, see \citeN{blanchet2009rare} and the references therein. \citeN{parekh1989quick} is probably one of the first to propose importance sampling methods with state-independent choices of importance distributions for single-station and tandem queues. \citeN{glasserman1995analysis} and \citeN{de2006analysis} further analyze the variance reduction performance of the Parekh-Walrand method, and prove that the state-independent importance sampling method in certain tandem queues is not asymptotically optimal. As a result, there has been a growing body of research related to designing importance sampling methods with  state-dependent choices of importance distributions for queueing systems, including single-class $G/G/1$ queue \shortcite{blanchet2007fluid}, single-class Jackson network \cite{dupuis2009importance}, two-class $M/M/1$ queue \cite{setayeshgar2011large}, and multi-class open/closed network \cite{dupuis2007subsolutions}. In these works, the importance distributions are generally derived from large deviation principles as well as differential game approaches, which  depend heavily on the specific system dynamics and are difficult to solve. 

Another stream of related literature lives in the off-policy estimation of long-horizon average reward in the field of reinforcement learning (RL), see \citeN{liu2001monte}, \shortciteN{xie2019towards} and the references therein. The Most common set of off-policy estimation methods is derived from importance sampling estimators. The likelihood weights are based on the product of the importance ratios of many steps in a trajectory \shortcite{liu2020understanding}, and thus variances in individual steps can accumulate multiplicatively. \shortciteN{liu2018breaking} shows that applying importance weighting on the state space rather than the trajectory space can substantially reduce estimation variance. Another set of off-policy estimation methods first fits a parametric model to learn the environment dynamics using data collected under the behavior policy, and then use this model to simulate trajectories under the evaluation policy \shortcite{chow2015robust}. These methods depend heavily on the model specification. A poor specification can result in poor model generalization and bias, even if the amount of data is infinite \shortcite{gottesman2019combining}.

\section{Problem Setup}
In the remaining part of the paper, we will illustrate the development of our proposed importance sampling algorithm and evaluate its performance using a 2-station tandem queueing network as our primary example. We emphasize here that extending our algorithm to other Markovian queueing networks is straightforward.
	\begin{figure}[htbp]
		\centering
		\includegraphics[scale=0.4]{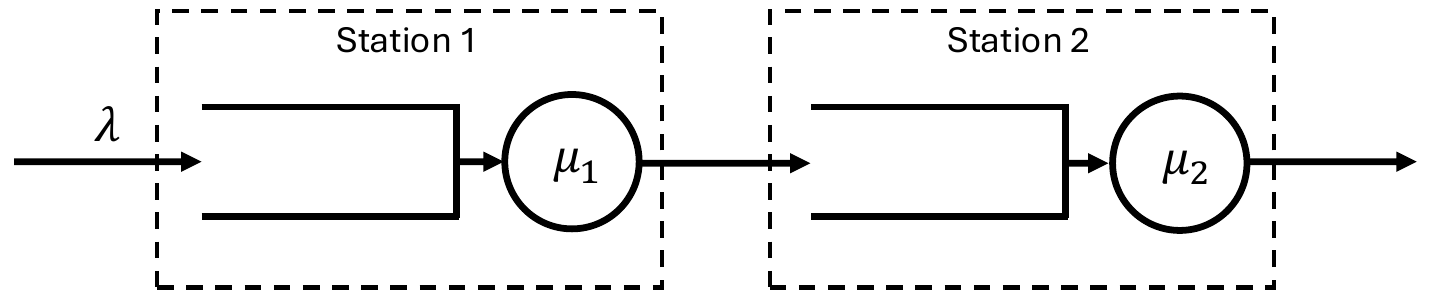}
		\caption{Structure of a 2-station tandem queueing network.} 
		\label{fig:tandem} 
	\end{figure} 

\subsection{Model \& Objective}\label{sec:setting}
	We consider a 2-station tandem queueing network, whose structure is shown in Figure \ref{fig:tandem}. Each job arrives according to a Poisson process with  rate $\lambda \in \mathbb{R}_+$. Each station $i \in \{1, 2\}$ has one single server which has i.i.d. exponential service times with rate $\mu_i \in \mathbb{R}_+ $. Consistent with real-world systems, we assume that the servers are stable, with parameters  satisfying
	\begin{align}\label{eq:stable}
		\lambda< \mu_2< \mu_1,
	\end{align} 
	and jobs are served in a first-come-first-serve manner. The arrival process and service times are assumed to be mutually independent. The system is empty initially.
	
	To evaluate this queueing system's service level performance, we particularly focus on the total number of jobs in the entire network.  In particular, the queueing dynamic with respect to the number $X_t^i$ of jobs in station $i$ at time $t$ could be written as
	\begin{equation*}
		\begin{aligned}
			X_t^1 &= A_t - D_t^1,\\
			X_t^2 &= D_t^1 - D_t^2,
		\end{aligned}
	\end{equation*}
	where $A_t$ is the Poisson arrival process and $D_t^i$ is the number of jobs leaving station $i$ up to time $t$. 
	
	The process $\{(X_t^1, X_t^2)\}$ forms a continuous time Markov chain (CTMC).  Under the stability assumption (\ref{eq:stable}), this CTMC admits a unique stationary distribution $\pi$. Let $X= (X^1, X^2)$ be a random vector following the distribution $\pi$. Our objective is to estimate a classic QoS metric in queue systems: the steady-state queue length overflow probability  \cite{whitt1993tail}, 
	\begin{align*}
		p_\gamma \equiv \mathbb{P}(X^1+X^2 \geq \gamma).
	\end{align*}
	The event $\{X^1+X^2 \geq \gamma\}$ becomes increasingly rare as the threshold $\gamma$ grows large, necessitating an efficient estimation algorithm. Before formally introducing  our algorithm, we first review the state-of-the-art  importance sampling method for stationary queue length overflow probabilities in tandem queueing networks. 
\subsection{Regenerative Importance Sampling Method}\label{sec:re_IS}
	The state-of-the-art importance sampling estimation for stationary tail probabilities relies on a regenerative simulation, see \citeN{asmussen2007stochastic} and \shortciteN{guang2022tail}.  For simplicity, we work with the uniformized discrete time Markov chains (DTMC) derived from the original tandem queueing network. We slightly abuse the notation by letting $\{(X_t^1, X_t^2)\}$ represent the  state of this uniformized DTMC. The transition kernel (probability) $P$ of this  uniformized  DTMC is summarized in the following equation,
	\begin{align*}
		P(X_{t+1}\mid X_t) = \begin{cases}
			\lambda/(\lambda + \mu_1 + \mu_2), &\text{if } X_{t+1}^1 =  X_{t}^1+1,\; X_{t+1}^2 =  X_{t}^2,\\
			\mu_1/(\lambda + \mu_1 + \mu_2), &\text{if } X_{t+1}^1 =  X_{t}^1 - 1,\; X_{t+1}^2 =  X_{t}^2+1,\\
			\mu_2/(\lambda + \mu_1 + \mu_2), &\text{if } X_{t+1}^1 =  X_{t}^1,\; X_{t+1}^2 =  X_{t}^2-1,\\
			1 - \sum_{\bx  \neq X_{t}} P(X_{t+1} = \bx \mid X_t), &\text{if } X_{t+1} = X_t.
		\end{cases}
	\end{align*}

	This uniformized  DTMC shares the same stationary distribution $\pi$ as the original tandem queueing network \cite{ross2014introduction}. The dynamics of  this DTMC $\{(X_t^1, X_t^2)\}$  can be viewed as a regenerative process, where the system regenerates whenever a job finds the system is empty upon departure. Let $\alpha$ denote the cycle length in units of time for a regenerative cycle. By the renewal reward theorem \cite{crane1977introduction}, the stationary tail probability can be expressed as
	\begin{align}\label{eq:prob_re}
		\mP(X^1+X^2 \geq \gamma) = \dfrac{\mE\left[\sum_{t=1}^{\alpha}\ones\{X_t^1+X_t^2 \geq \gamma\}\right]}{\mE[\alpha]}.
	\end{align}
	Thus, estimating this probability reduces to separately estimating the numerator  and the denominator.
	
	We first describe the importance sampling estimation of the numerator in (\ref{eq:prob_re}) using data from an alternative system.  In each regenerative cycle, the alternative system first follows the dynamic of the uniformized DTMC derived from a modified tandem queueing network with exponential arrival rate $\tilde{\lambda}$ and exponential service rates $\tilde{\mu}_i$. This modified system  is intentionally unstable with 
	$$\tilde{\lambda}\geq \tilde{\mu}_1, \quad \tilde{\lambda}\geq \tilde{\mu}_2,$$
	so that  the target event $\{X_t^1+X_t^2 \geq \gamma\}$ can be observed frequently. Once the total number of jobs in the system reaches $\gamma$ at time $\tilde{\tau}$, the dynamic of the  system is switched back to the uniformized DTMC derived from the original tandem queueing network. Let $\tX_t^i$ denote the number of jobs in station $i$ at time $t$ in the alternative system, and $\tilde{\alpha}$ be the regenerative cycle length. Then, we have the following equality,
	\begin{align*}
		\mE\left[\sum_{t=1}^{\alpha}\ones\{X_t^1+X_t^2 \geq \gamma\}\right] = \mE\left[\left(\sum_{t=1}^{\tilde\alpha}\ones\{\tX_t^1+\tX_t^2 \geq \gamma\}\right)\cdot\left(\prod_{t=1}^{\tilde \tau}\dfrac{P(\tX_{t}\mid \tX_{t-1})}{Q(\tX_{t}\mid \tX_{t-1})}\right)\right],
	\end{align*}
	where  $P$ and $Q$ are the transition kernels of the uniformized DTMC derived from the original and alternative queueing networks.
	
	Since the the denominator in (\ref{eq:prob_re}) does not involve a rare event, it is directly estimated using  data  from the original system. Let us  independently generate $m_1$ and $m_2$ regenerative cycles of data from the alternative and original systems respectively. Using an extra subscript $i$ to denote the $i$th cycle, the  state-of-the-art  importance sampling estimator is given by 
	\begin{align*}
		\hat{\bP}(X^1+X^2 \geq \gamma) = \dfrac{m_1^{-1}\sum_{i=1}^{m_1} \left[\left(\sum_{t=1}^{\tilde\alpha_i}\ones\{\tX_{t,i}^1+\tX_{t,i}^2 \geq \gamma\}\right)\cdot\left(\prod_{t=1}^{\tilde \tau_i}\frac{P(\tX_{t,i}\mid \tX_{t-1,i})}{Q(\tX_{t,i}\mid \tX_{t-1,i})}\right)\right]}{m_2^{-1}\sum_{i=1}^{m_2}\alpha_i}.
	\end{align*}

	The performance of this importance sampling estimator hinges on the choice of parameters $(\tilde{\lambda}, \tilde{\mu}_1,  \tilde{\mu}_2)$ for the alternative system.  \citeN{de2006analysis} proves that letting  $(\tilde{\lambda}, \tilde{\mu}_1,  \tilde{\mu}_2) = (\mu_2, \mu_1, \lambda)$  is the only possible choice  so that the IS estimator can use the orderwisely smallest amount of data  to achieve the same level of variance, which is the so-called \textit{asymptotically efficient} estimator. However, as implied in \citeN{glasserman1995analysis},  this IS estimator with any choice of parameters $(\tilde{\lambda}, \tilde{\mu}_1,  \tilde{\mu}_2)$ is not asymptotically  efficient  when the original system parameters satisfy
	\begin{align}\label{eq:optimal_cond}
		\dfrac{\mu_2(\mu_1+\mu_2)}{(\lambda+\mu_1)^2} > 1.
	\end{align}
	
\begin{figure}[htb]
		\centering
		\includegraphics[scale=0.45]{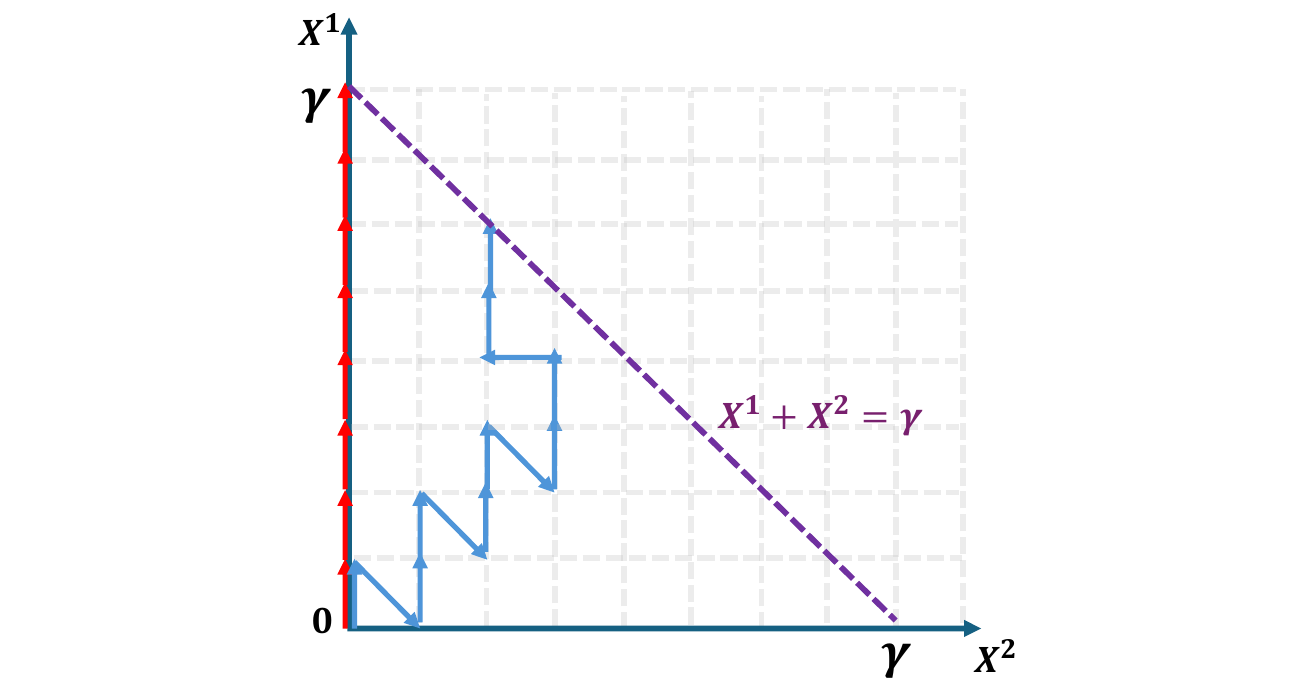}
		\caption{Two representative sample paths reaching the  target event $\{X^1 + X^2 \geq \gamma\}$ from the empty state.} 
		\label{fig:sample_path} 
	\end{figure}

	The IS estimator suffers from the excessive variance introduced by the path-dependent likelihood ratio $\prod_{t=1}^{\tilde \tau_i}\frac{P(\tX_{t,i}\mid \tX_{t-1,i})}{Q(\tX_{t,i}\mid \tX_{t-1,i})}$. The main reason is that  there are a wide range of sample paths in a regenerative cycle that can reach the target event $\{X^1_t + X^2_t \geq \gamma\}$, see Figure \ref{fig:sample_path}, and using a certain alternative system with parameters $(\tilde{\lambda}, \tilde{\mu}_1,  \tilde{\mu}_2)$ may not reduce the variance contributed by the path-dependent likelihood ratio on every sample path. For instance, using the alternative system with parameters $(\tilde{\lambda}, \tilde{\mu}_1,  \tilde{\mu}_2) = (\mu_2, \mu_1, \lambda)$ could reduce the variance contributed by the path-dependent likelihood ratio on the blue path in Figure \ref{fig:sample_path}, however, the likelihood ratio on the red path will then introduce an explosive amount of variance to the IS estimator \cite{glasserman1995analysis}.  This limitation motivates us to propose a robust IS method that can achieve variance reduction with a wider choice of alternative systems.

\newpage
\section{Importance Sampling with Marginal Stationary Distribution}\label{sec:alg}
	Inspired by \shortciteN{liu2018breaking}, we intend to apply the importance sampling directly on the ``marginal'' stationary distribution of each sample data. Specifically, let us consider an alternative 2-node tandem queueing network with exponential arrival rate $\tilde{\lambda}$ and exponential service rates $\tilde{\mu}_i$, where the parameters satisfy the stability condition
	\[\tilde{\lambda} < \tilde{\mu}_2 \leq  \tilde{\mu}_1.\]
	This queueing network has  a stationary distribution $\tilde{\pi}$. Let $\tilde{X} = (\tilde X^1, \tilde X^2)$ be a random vector following the distribution $\tilde \pi$. The  steady-state queue length overflow probability for the original queueing network can then be expressed as 
	\begin{align*}
		\mathbb{P}(X^1+X^2 \geq \gamma) &= \sum_{x_1,x_2} \ones\{x_1 + x_2 \geq \gamma\}\cdot \pi(x_1, x_2),\\
		&= \sum_{x_1,x_2} \ones\{x_1 + x_2 \geq \gamma\}\dfrac{\pi(x_1, x_2)}{\tilde \pi(x_1, x_2)}\cdot \tilde \pi(x_1, x_2),\\
		&= \mE_{\tX \sim \tpi} \left[ \ones\{\tX^1 + \tX^2 \geq \gamma\}\dfrac{\pi(\tX)}{\tpi(\tX)} \right].
	\end{align*}
	It is noted that $\ones\{\tX^1 + \tX^2\geq \gamma\}\frac{\pi(\tX)}{\tpi(\tX)}$ is a promising choice of the importance sampling estimator, as it avoids the excessive variance introduced by the likelihood ratio that depends on the whole sample path. However,
	a critical challenge occurs: the stationary distributions $\pi$ and $\tpi$ are typically unknown for complex queueing networks. Therefore, a vital step in our proposed algorithm is to learn the stationary likelihood ratio $\pi/\tilde{\pi}$, which is illustrated in detail in the follow section.

	\subsection{Learning Stationary Likelihood Ratio $\pi/\tilde{\pi}$} 
	Let $P$ and $Q$ be the transition kernels of the uniformized discrete time Markov chain (DTMC) derived from the original and alternative tandem queueing networks. Suppose $X$ and  $\tX$ are sampled from $\pi$ and $\tpi$, respectively, and  $\tX^{\prime}$ is sampled from $Q(\cdot\mid \tX)$, which also follows the distribution $\tpi$. By the definition of Markov chain's transition kernel and stationary distribution, we can derive the the following equality for the expectation of $f(X)$ for any function $f$:
	\begin{align*}
		\mE_{X \sim \pi}[f(X)] = \mE\left[f(\tX^{\prime}) \cdot \dfrac{\pi( \tX^{\prime})}{\tilde{\pi}( \tX^{\prime})}\right]= \mE\left[f(\tX^{\prime}) \cdot \dfrac{\pi( \tX)}{\tilde{\pi}( \tX)}\cdot \dfrac{P( \tX^{\prime} |  \tX)}{ Q( \tX^{\prime} |  \tX)}\right].
	\end{align*}
	This observation can be leveraged to obtain the following key property of the stationary likelihood ratio $\pi/\tilde{\pi}$.
	\begin{proposition}\label{prop:stationaryRatioEst}\shortcite{liu2018breaking}
		A function $w(x)$ equals $\pi(x)/\tpi(x)$ (up to a constant factor) if and only if it satisfies
		\begin{equation}\label{eq:stationaryRatioEst}
			\begin{aligned}
				\mE [f(\tX')\Delta(w; \tX, \tX^{\prime})] &= 0, \quad \text{for any function }  f,\\
				\text{with} \quad \Delta(w; \tX, \tX^{\prime}) &\equiv w(\tX)\cdot\dfrac{P(\tX^{\prime}\mid \tX)}{Q(\tX^{\prime}\mid \tX)} - w(\tX^{\prime}).
			\end{aligned}
		\end{equation}
	\end{proposition}
	Following the approach in \shortciteN{liu2018breaking}, we can estimate $ \pi/\tpi$ by solving the following min-max problem:
	\begin{align*}
		\min_{w\neq 0}  \; L(w) \equiv \max_{f \in \cF}  \mE [f(\tX')\Delta(w; \tX, \tX^{\prime})]^2,
	\end{align*}
	where the condition $w\neq 0$ is added to avoid the trivial solution $w \equiv 0$, and $\cF$ is a set of test functions that should be rich enough to identify $w$.  Let $\cH$ be a reproducing kernel Hilbert space (RKHS) of functions with a positive deﬁnite kernel $k(r, \bar{r})$. A good choice of $\cF$ is the unit ball of the RKHS $\cH$, i.e., $\cF \equiv \{f \in \cH: ||f||_{\cH} \leq 1\}$,  as it yields a closed form representation of the loss function $L(w)$:
	\begin{align*}
		L(w) = \mE[\Delta(w; \tX_a, \tX_a^{\prime})\Delta(w; \tX_b, \tX_b^{\prime})k(\tX_a^{\prime}, \tX_b^{\prime})].
	\end{align*}
	Here, $(\tX_a, \tX_a^{\prime})$ and  $(\tX_b, \tX_b^{\prime})$ are independent transition pairs, where $\tX_i $ and $\tX_i^{\prime}$ follow $\tpi$  and $Q(\cdot\mid \tX_i )$ respectively, for $i \in \{a,b\}$. See \citeN{berlinet2011reproducing} for additional background on RKHS.
	
	\subsection{A Machine Learning Based Importance Sampling Algorithm (MLIS)}
	According to Proposition \ref{prop:stationaryRatioEst}, a function $w$ is proportional to the stationary likelihood ratio $\pi/\tilde{\pi}$ if and only if the loss function $L(w) = 0$. Following this idea, we parameterize $w(x) = w_\theta(x)$  as a neural network. This  neural network is then trained by minimizing the loss function $L(w_\theta)$ using a stochastic gradient method such as Adam. 
	
	In practice, we design a robust loss function $L(w)$ for a better performance of our algorithm, namely,
	\begin{align*}
		L(w) = \mE\left[\mE[\Delta(w; \tX_a, \tX_a^{\prime})\Delta(w; \tX_b, \tX_b^{\prime})k(\tX_a^{\prime}, \tX_b^{\prime}) \mid k = k_i]\right]+ \dfrac{\alpha}{2}(\mE[w(\tX_a)] - 1)^2.
	\end{align*} 
	where $(\tX_a, \tX_a^{\prime})$ and  $(\tX_b, \tX_b^{\prime})$ are independent transition pairs,  $\tX_i $ and $\tX_i^{\prime}$ follow $\tpi$  and $Q(\cdot\mid \tX_i )$ respectively, for $i \in \{a,b\}$, and  the regularization parameter $\alpha > 0$.  In detail, we enlarge the set of test function by assuming that the test function has an equal possibility to lay in a range of different reproducing kernel Hilbert spaces  with kernel functions $k_i$. The extra  regularization term is added in  $L(w)$ to  prevent the neural network $w_{\theta}$ from converging to the trivial estimation $w_{\theta} \equiv 0$.

	In conclusion, we summarize our machine learning based steady state importance sampling algorithm (MLIS) in Algorithm \ref{alg:1}.

	\algnewcommand\algorithmicinput{\textbf{Input:}}
	\algnewcommand\Input{\item[\algorithmicinput]}
	\algnewcommand\algorithmicinitial{\textbf{Initiate}}
	\algnewcommand\Initial{\item[\algorithmicinitial]}
	\algnewcommand\algorithmicoutput{\textbf{Output:}}
	\algnewcommand\Output{\item[\algorithmicoutput]}

	\begin{algorithm}[ht]
		\caption{Machine Learning Based  Importance Sampling (MLIS)}\label{alg:1}
		\begin{algorithmic}
			\Input Simulation data $\cD = \{X_t\}_{t=1}^T$ of the uniformized DTMC derived from the alternative system. Regularization parameter $\alpha$. Kernel function set $\{k_i(\bx, \by)\}$.
			\Initial the density ratio $w(r) = w_{\theta}(r)$ to be a neural network parameterized by $\theta$.
			\For{$\text{iteration } = 1, 2, ...$}
			\State Randomly choose a batch $\cM_D$ of size $b_D$ from the data $\cD$, i.e., $\cM \subset \{1, ..., T-1\}$.
			\State Randomly choose a batch $\cM_k$ of size $b_k$ from the kernel function set $\{k_i(\bx, \by)\}$.
			\State \textbf{Compute}  the sample average loss function
			\begin{align*}
				\hat{L}(w_{\theta}) &= \dfrac{1}{b_D^2b_k}\sum_{i,j\in \cM_D}\sum_{u \in \cM_k} \Delta(w_\theta; X_i, X_{i+1})\Delta(w_\theta; X_j, X_{j+1})k_{u}(X_{i+1}, X_{j+1}) \\
				&\quad+ \dfrac{\alpha}{2}\left(\dfrac{1}{b_D}\sum_{i \in \cM_D}w_{\theta}(X_i) - 1\right)^2.
			\end{align*}
			\State \textbf{Update} the parameter $\theta$ by $\theta \leftarrow \theta - \epsilon \nabla_{\theta}\hat{L}(w_{\theta})$.

			\EndFor
			\Output Estimate the tail probability of the original system by 
			\begin{align*}
				\hat{\bP}_{MLIS}(X^1+X^2 \geq \gamma) = \dfrac{\sum_{t=1}^{T}w_{\theta}(X_t)\cdot \ones\{X_t^1+X_t^2 \geq \gamma\}}{\sum_{t=1}^{T}w_{\theta}(X_t)}.
			\end{align*}
		\end{algorithmic}
	\end{algorithm}

	\section{Numerical Experiments} \label{sec:experiment}
	In this section, we would compare the performance of our Algorithm \ref{alg:1} with different benchmark methods for a 2-node tandem queueing network with parameters $(\lambda, \mu_1, \mu_2)$ satisfying the inequality in (\ref{eq:optimal_cond}). It is well known that the queue length overflow probability at steady state for a tandem queueing network has an explicit expression, i.e.,
	\[\mathbb{P}(X^1+X^2 \geq \gamma) = \dfrac{(1-\rho_1)\rho_2^{\gamma+1} - (1-\rho_2)\rho_1^{\gamma+1}}{\rho_2 - \rho_1},\]
	where the load parameter $\rho_i = \lambda/\mu_i$ for $i\in \{1,2\}.$ 
	
	\subsection{Benchmarks and Performance Metrics}
	The first benchmark method is the marginal importance sampling method (MIS). Let us consider an alternative 2-node tandem queueing network with exponential arrival rate $\tilde{\lambda}$ and exponential service rates $\tilde{\mu}_i$, where the parameters satisfy the stability condition $\tilde{\lambda} < \tilde{\mu}_2 \leq  \tilde{\mu}_1$ and queueing network has  a stationary distribution $\tilde{\pi}$. Let us generate a sample path $\{\tX_t\}$ of the uniformized DTMC derived from this  alternative system, then the MIS estimator is computed by the following equation, i.e.,
	\begin{align*}
		\hat{\bP}_{MIS}(X^1+X^2 \geq \gamma) = \dfrac{\sum_{t=1}^{T}\frac{\pi}{\tilde{\pi}}(\tX_t)\cdot \ones\{\tX_t^1+\tX_t^2 \geq \gamma\}}{\sum_{t=1}^{T}\frac{\pi}{\tilde{\pi}}(\tX_t)}.
	\end{align*}
	The key distinction between MIS estimator and our Algorithm \ref{alg:1} is that MIS estimator directly uses the stationary likelihood ratio $\pi/\tpi$, whereas our Algorithm \ref{alg:1} learns this ratio via a neural network. While MIS estimator is expected to outperform our algorithm,  we hope the performance gap, that is the price of learning the stationary likelihood ratio $\pi/\tpi$, would be small.
	
	The other benchmark method is the regenerative importance sampling method (RIS) discussed in Section \ref{sec:re_IS}, where the RIS estimator is computed by the following equation, i.e.,
	\begin{align*}
		\hat{\bP}_{RIS}(X^1+X^2 \geq \gamma) = \dfrac{m_1^{-1}\sum_{i=1}^{m_1} \left[\left(\sum_{t=1}^{\tilde\alpha_i}\ones\{\tX_{t,i}^1+\tX_{t,i}^2 \geq \gamma\}\right)\cdot\left(\prod_{t=1}^{\tilde \tau_i}\frac{P(\tX_{t,i}\mid \tX_{t-1,i})}{Q(\tX_{t,i}\mid \tX_{t-1,i})}\right)\right]}{m_2^{-1}\sum_{i=1}^{m_2}\alpha_i}.
	\end{align*}
	Specifically, the numerator is computed by $m_1$ regenerative cycles of data from the alternative system with parameters $(\tilde{\lambda}, \tilde{\mu}_1,  \tilde{\mu}_2) = (\mu_2, \mu_1, \lambda)$, and the denominator is  computed by $m_2$ regenerative cycles of data from the original system. For a fair comparison, we let the number of data sample generated for computing a RIS estimator be equal to (or very closed to) that generated for computing a MIS estimator, namely
	\[\sum_{i=1}^{m_1} \tilde\alpha_i + \sum_{i=1}^{m_2}\alpha_i \approx T.\]
	
	The mean squared error (MSE) for each importance sampling method is estimated based on 500 independent rounds of simulation, that is
	\[
	MSE_{\gamma} = \dfrac{1}{500}\sum_{i=1}^{500}\left(\hat{\bP}_{a, i}(X^1+X^2 \geq \gamma) - \mathbb{P}(X^1+X^2 \geq \gamma)\right)^2,
	\]
	where $\hat{\bP}_{a, i}(X^1+X^2 \geq \gamma)$ is the IS estimator for some method $a$ in the $i$-th round of simulation. We evaluate the performance of different IS methods in term of their relative mean squared errors (rMSE), namely
	\begin{align*}
		rMSE_{\gamma}  = \dfrac{\sqrt{MSE_{\gamma} }}{\mathbb{P}(X^1+X^2 \geq \gamma)}.
	\end{align*}

	\subsection{Implementation Details for Algorithm \ref{alg:1}}\label{sec:imple_alg}
	We hereby reveal the implementation details for our machine learning based importance sampling method (MLIS). Our algorithm uses sample data of the uniformized DTMC derived from the alternative tandem queueing network with parameters $(\tilde{\lambda}, \tilde{\mu}_1, \tilde{\mu}_2)$, which satisfy the stability condition $\tilde{\lambda} < \tilde{\mu}_2 \leq  \tilde{\mu}_1$.
	
	The performance of our algorithm is evaluated with different choices of alternative systems. For each alternative system,  a fully connected neural network $w_{\theta}$ is constructed to approximate the stationary likelihood ratio $\pi/\tilde{\pi}$ of  queue length distributions in the original and alternative systems. Each neural network consists of two hidden layers, each with $1,024$ neurons activated by the ReLU function, and the output layer is activated by the softplus function, i.e., $\text{softplus}(x) = \log(1+\exp(x))$, to ensure a nonnegative result.  
	
	Four types of kernel functions are used to construct the loss function for training the neural network. Namely, for $\sigma > 0$ and $\bx, \by \in \mR^n$,
	\begin{alignat*}{2}
		\text{Gaussian:}  &\quad k_{GS}(\bx, \by) &&= \exp\left(-\dfrac{\|\bx - \by\|^2}{2\sigma^2}\right), \\
		\text{Laplacian:} &\quad k_{LP}(\bx, \by)      &&= \exp\left(-\dfrac{\|\bx - \by\|}{\sigma}\right),\\
		\text{Inverse multiquadratic:} &\quad k_{IM}(\bx, \by) &&= \left(1+\dfrac{||\bx - \by||^2}{\sigma^2}\right)^{-1/2},\\
		\text{Linear:} &\quad k_{LN}(\bx, \by) &&= \bx^T \by.
	\end{alignat*}
	The parameter $\sigma$ is chosen to be the median of Euclidean distances computed using  a batch of sample data. We set the data batch size $b_D = 3,000$, kernel batch size $b_k = 4$.

	\subsection{Performance Comparison}
	The simulation is implemented in Python and runs on a 2021 version MacBook Pro with an 8-core Apple M1 pro chip. The numerical results and computation times are reported in Table \ref{table:performance_comparison_between_methods_a}. It is clear that our algorithms (MLIS) with different choices of alternative systems all achieve  lower relative MSE than the state-of-the-art RIS  method. Our algorithms  (MLIS) maintain a low level of relative MSE even when the target event $\{X^1+X^2 \geq \gamma\}$ becomes rarer. Conversely, the relative MSE for the state-of-the-art  RIS method increases dramatically. This result is consistent with our previous  discussion in Section \ref{sec:re_IS} since a wider range of sample paths in a regenerative cycle can reach the target event when the threshold parameter $\gamma$ increases. Moreover, our algorithms  (MLIS) with different choices of alternative systems maintain relative MSE within the same order of magnitude as the MIS method, which indicates that our algorithms estimate the true stationary likelihood ratio $\pi/\tilde{\pi}$ accurately.

	\begin{table}[ht]
		\centering
		\renewcommand{\arraystretch}{1.3}
		\setlength{\tabcolsep}{8pt}
        \caption{Performance of different importance sampling estimation methods in a two-node tandem queueing network with parameter $(\lambda, \mu_1, \mu_2) = (1/10, 23/50, 11/25)$. The relative MSE is estimated based on 500 independent rounds of simulation. Time horizon $T = 100,000$.} 
		\begin{tabular}{|l|cccccc|}
			\hline
			\multicolumn{1}{|c|}{Target Probability}                                                                     & Method & $\tilde{\lambda}$     & $\tilde{\mu}_1$       & $\tilde{\mu}_2$       & rMSE  & \begin{tabular}[c]{@{}c@{}}Running\\ Time (s)\end{tabular} \\ \hline
			\multirow{7}{*}{\begin{tabular}[c]{@{}l@{}}$\mP(X^1+X^2 \geq 18)$\\ $ = 4.891 \times 10^{-10}$\end{tabular}} & MLIS   & \multirow{2}{*}{3/11} & \multirow{2}{*}{4/11} & \multirow{2}{*}{4/11} & 0.325 & 611                                                        \\
			& MIS    &                       &                       &                       & 0.153 & 417                                                        \\ \cline{2-7} 
			& MLIS   & \multirow{2}{*}{5/17} & \multirow{2}{*}{6/17} & \multirow{2}{*}{6/17} & 0.278 & 598                                                        \\
			& MIS    &                       &                       &                       & 0.168 & 432                                                        \\ \cline{2-7} 
			& MLIS   & \multirow{2}{*}{7/23} & \multirow{2}{*}{8/23} & \multirow{2}{*}{8/23} & 0.191 & 603                                                        \\
			& MIS    &                       &                       &                       & 0.145 & 409                                                        \\ \cline{2-7} 
			& RIS    & 11/25                 & 23/50                 & 1/10                  & 0.573 & 698                                                        \\ \hline
			\multirow{7}{*}{\begin{tabular}[c]{@{}l@{}}$\mP(X^1+X^2 \geq 18)$\\ $ = 2.712 \times 10^{-11}$\end{tabular}} & MLIS   & \multirow{2}{*}{3/11}     & \multirow{2}{*}{4/11} & \multirow{2}{*}{4/11} & 0.311 & 609                                                        \\
			& MIS    &                       &                       &                       & 0.195 & 441                                                        \\ \cline{2-7} 
			& MLIS   & \multirow{2}{*}{5/17} & \multirow{2}{*}{6/17} & \multirow{2}{*}{6/17} & 0.236 & 612                                                        \\
			& MIS    &                       &                       &                       & 0.187 & 428                                                        \\ \cline{2-7} 
			& MLIS   & \multirow{2}{*}{7/23} & \multirow{2}{*}{8/23} & \multirow{2}{*}{8/23} & 0.227 & 607                                                        \\
			& MIS    &                       &                       &                       & 0.158 & 425                                                        \\ \cline{2-7} 
			& RIS    & 11/25                 & 23/50                 & 1/10                  & 1.859 & 701                                                        \\ \hline
			\multirow{7}{*}{\begin{tabular}[c]{@{}l@{}}$\mP(X^1+X^2 \geq 20)$\\ $ = 1.489 \times 10^{-12}$\end{tabular}} & MLIS   & \multirow{2}{*}{3/11} & \multirow{2}{*}{4/11} & \multirow{2}{*}{4/11} & 0.467 & 597                                                        \\
			& MIS    &                       &                       &                       & 0.259 & 437                                                        \\ \cline{2-7} 
			& MLIS   & \multirow{2}{*}{5/17} & \multirow{2}{*}{6/17} & \multirow{2}{*}{6/17} & 0.464 & 603                                                        \\
			& MIS    &                       &                       &                       & 0.202 & 431                                                        \\ \cline{2-7} 
			& MLIS   & \multirow{2}{*}{7/23} & \multirow{2}{*}{8/23} & \multirow{2}{*}{8/23} & 0.453 & 610                                                        \\
			& MIS    &                       &                       &                       & 0.171 & 447                                                        \\ \cline{2-7} 
			& RIS    & 11/25                 & 23/50                 & 1/10                  & 3.241 & 702                                                        \\ \hline
		\end{tabular}
\label{table:performance_comparison_between_methods_a}
	\end{table}

	\subsection{Influence of Kernel Functions}
	In this section, we evaluate the performance of our Algorithm \ref{alg:1} with a total of 15 different combinatorial choices of the  four  commonly used kernel functions mentioned in Section \ref{sec:imple_alg}. We set the data batch size $b_D = 3,000$ and the kernel batch size $b_k$ to equal the number of selected kernels in each experiment.

	The numerical results and training times are reported in Table \ref{table:performance_comparison_between_kernels}. A key observation is that the linear kernel plays an important role in improving the performance of our algorithm, and any set of kernel functions without linear kernel cannot help our algorithm learn the correct stationary likelihood ratio $\pi/\tilde{\pi}$. Moreover, although using more kernels can only marginally reduce our estimator's  relative mean squared  error, it significantly improves the efficiency of  learning the likelihood ratio.
	
	\begin{table}[ht]
		\centering
		\renewcommand{\arraystretch}{1.5}
		\setlength{\tabcolsep}{3pt}
        \caption{Performance of  Algorithm \ref{alg:1} with different choices of kernel sets in a two-node tandem queueing network with parameter $(\lambda, \mu_1, \mu_2) = (1/10, 23/50, 11/25)$. The alternative tandem queueing network  is chosen with  parameter $(\tilde{\lambda}, \tilde{\mu}_1, \tilde{\mu}_2) = (7/23, 8/23, 8/23)$. The relative MSE is estimated based on 500 independent rounds of simulation. Time horizon $T = 100,000$.} 
		\begin{tabular}{|c|c|c|c|c|}
			\hline
			Chosen Kernels                                                                        & \begin{tabular}[c]{@{}c@{}}rMSE for Estimating\\ $\mP(X^1+X^2 \geq 16)$\end{tabular} & \begin{tabular}[c]{@{}c@{}}rMSE for Estimating\\ $\mP(X^1+X^2 \geq 18)$\end{tabular} & \begin{tabular}[c]{@{}c@{}}rMSE for Estimating\\ $\mP(X^1+X^2 \geq 20)$\end{tabular} & \begin{tabular}[c]{@{}c@{}}Training\\ Time (s)\end{tabular} \\ \hline
			$k_{LN}$                                                                                &                         1.235                                                             &             2.811                                                                         &              5.678                                                                        & 2314                                                        \\ \hline
			$k_{LN}$, $k_{GS}$                                                                     & 0.254                                                                                & 0.288                                                                                & 0.777                                                                                & 2502                                                        \\ \hline
			$k_{LN}$, $k_{LP}$                                                                    & 0.324                                                                                & 0.253                                                                                & 0.677                                                                                & 2030                                                        \\ \hline
			$k_{LN}$, $k_{IM}$                                                                      & 0.226                                                                                & 0.327                                                                                & 0.773                                                                                & 1901                                                        \\ \hline
			$k_{LN}$, $k_{GS}$, $k_{LP}$          & 0.265                                                                                & 0.295                                                                                & 0.825                                                                                & 1561                                                        \\ \hline
			$k_{LN}$, $k_{GS}$, $k_{IM}$               & 0.245                                                                                & 0.337                                                                                & 0.904                                                                                & 1493                                                        \\ \hline
			$k_{LN}$, $k_{LP}$, $k_{IM}$           & 0.247                                                                                & 0.342                                                                                & 0.562                                                                                & 1447                                                        \\ \hline
			$k_{LN}$,  $k_{GS}$, $k_{LP}$, $k_{IM}$     & 0.191                                                                                & 0.227                                                                                & 0.453                                                                                & 982                                                         \\ \hline\hline
			MIS                                                                  &  0.145                                                                        &  0.158                                                                        &  0.171                                                                        &  -
			\\ \hline
		\end{tabular}
        \label{table:performance_comparison_between_kernels}
	\end{table}

	\section{Conclusion}\label{sec:conclusion}
	In this paper, we propose a novel algorithm to estimate the tail probability of the stationary distribution of queuing networks, combining importance sampling with machine learning techniques. In detail, our algorithm applies importance sampling directly on the stationary distributions, to avoid the excessive variance encountered by the classic path-dependent  method, and leverages machine learning techniques to approximate the likelihood ratio corresponding to the stationary distributions. Numerical experiments demonstrate that our algorithm, across a reasonable wide range of importance distributions, consistently outperforms the benchmark methods. 
	
	There are several interesting paths for future exploration. First, despite the numerical findings showing that our algorithm's performance is stable across a reasonably wide range of importance distributions, it remains worthwhile to determine,  through either analytical or numerical means, the optimal set of importance distributions to further reduce the variance. Additionally, our numerical results suggest that the linear kernel has a significant impact on enhancing our algorithm's performance. We hypothesize that the linear kernel effectively captures the tail structure of the true stationary likelihood ratio $\pi/\tpi$, but a formal mathematical justification remains necessary. Such analysis could deepen the theoretical understanding of our algorithm.

\footnotesize

\bibliographystyle{wsc}

\bibliography{bibliography}

\end{document}